\documentclass[twocolumn]{IEEEtran}
\usepackage{color}
\ifCLASSOPTIONcompsoc
  \usepackage[nocompress]{cite}
\else
  \usepackage{cite}
\fi
\usepackage{multirow}
\usepackage{pifont}
\newcommand{\cmark}{\ding{51}}%
\newcommand{\xmark}{\ding{55}}%

\ifCLASSINFOpdf
  \usepackage[pdftex]{graphicx}
  \graphicspath{{./figs/}{./final_version/photos/}}
  \DeclareGraphicsExtensions{.pdf,.jpeg,.jpg,.png}
\else
\fi
\usepackage{url}

\begin{document}
%
\title{T-CNN: Tubelets with Convolutional Neural Networks for Object Detection from Videos}
\author{Kai~Kang*, Hongsheng~Li*, Junjie~Yan, Xingyu~Zeng, Bin~Yang, Tong~Xiao, Cong~Zhang, Zhe~Wang, Ruohui~Wang, Xiaogang~Wang,~\IEEEmembership{Member,~IEEE,}, and Wanli~Ouyang,~\IEEEmembership{Senior~Member,~IEEE,}
\thanks{Copyright © 2017 IEEE. Personal use of this material is permitted. However, permission to use this material for any other purposes must be obtained from the IEEE by sending an email to pubs-permissions@ieee.org.}
\thanks{This work is supported in part by SenseTime Group Limited, in part by the General Research Fund through the Research Grants Council of Hong Kong under Grants CUHK14213616, CUHK14206114, CUHK14205615, CUHK419412, CUHK14203015, CUHK14239816, CUHK14207814, in part by the Hong Kong Innovation and Technology Support Programme Grant ITS/121/15FX, in part by the China Postdoctoral Science Foundation under Grant 2014M552339, in part by National Natural Science Foundation of China (No. 61371192), and in part by ONR N00014-15-1-2356.}
\thanks{*Kai Kang and Hongsheng Li share co-first authorship.}
\thanks{Wanli Ouyang is the corresponding author. (wlouyang@ee.cuhk.edu.hk)}
\thanks{Kai Kang, Hongsheng Li, Tong Xiao, Zhe Wang, Ruohui Wang, Xiaogang Wang, and Wanli Ouyang are with The Chinese University of Hong Kong.}
\thanks{Cong Zhang is with Shanghai Jiao Tong University, China.}
\thanks{Junjie Yan, Xingyu Zeng are with the SenseTime Group Limited.}
\thanks{Bin Yang is with the Computer Science Department, University of Toronto.}

}
\IEEEtitleabstractindextext{%
\begin{abstract}
The state-of-the-art performance for object detection has been significantly improved over the past two years.
Besides the introduction of powerful deep neural networks such as GoogleNet \cite{googlenet} and VGG \cite{vgg2014simonyan}, novel object detection frameworks such as R-CNN \cite{girshick2014rich} and its successors, Fast R-CNN \cite{girshick2015fast} and Faster R-CNN \cite{ren2015faster}, play an essential role in improving the state-of-the-art.
Despite their effectiveness on still images, those frameworks are not specifically designed for object detection from videos.
Temporal and contextual information of videos are not fully investigated and utilized.
In this work, we propose a deep learning framework that incorporates temporal and contextual information from tubelets obtained in videos, which dramatically improves the baseline performance of existing still-image detection frameworks when they are applied to videos. It is called T-CNN, i.e. tubelets with convolutional neueral networks. The proposed framework won newly introduced object-detection-from-video (VID) task with provided data in the ImageNet Large-Scale Visual Recognition Challenge 2015 (ILSVRC 2015). Code is publicly available at \url{https://github.com/myfavouritekk/T-CNN}.

\end{abstract}

} 

\maketitle

\IEEEdisplaynontitleabstractindextext

%
\IEEEpeerreviewmaketitle

\section{Introduction}
\label{sec:introduction}
\IEEEPARstart{I}{n} the last several years, the performance of object detection has been significantly improved with the success of novel deep convolutional neural networks (CNN) \cite{he2016deep,googlenet,vgg2014simonyan,ioffe2015batch} and object detection frameworks \cite{girshick2014rich,ouyang2015deepid,girshick2015fast,ren2015faster}.
The state-of-the-art frameworks for object detection such as R-CNN \cite{girshick2014rich} and its successors \cite{girshick2015fast,ren2015faster} extract deep convolutional features from region proposals and classify the proposals into different classes.
DeepID-Net \cite{ouyang2015deepid} improved R-CNN by introducing box pre-training, cascading on region proposals, deformation layers and context representations.
Recently, ImageNet introduces a new challenge for object detection from videos (VID), which brings object detection into the video domain.
In this challenge, an object detection system is required to automatically annotate every object in $30$ classes with its bounding box and class label in each frame of the videos, while test videos have no extra information pre-assigned, such as user tags. VID has a broad range of applications on video analysis.

\begin{figure}[tb]
    \centering
    \includegraphics[width=\linewidth]{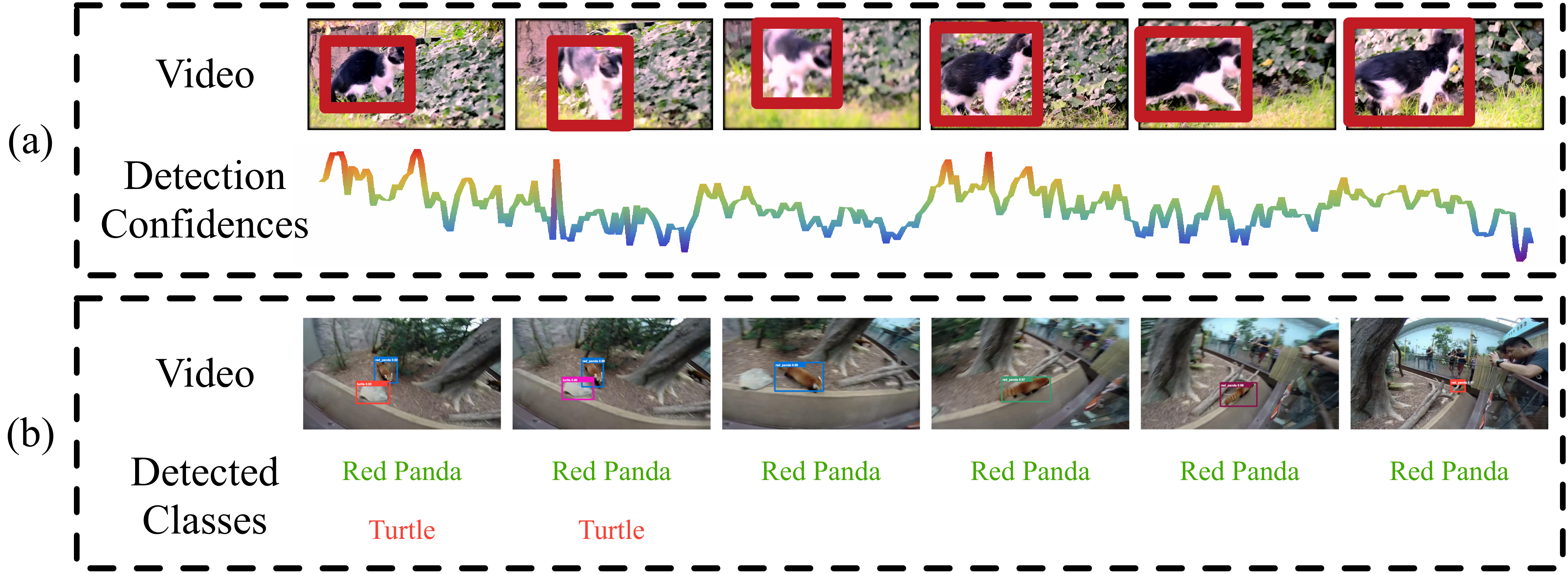}
    \caption{Limitations of still-image detectors on videos. (a) Detections from still-image detectors contain large temporal fluctuations, because they do not incorporate temporal consistency and constraints. (b) Still-image detectors may generate false positives solely based the information on single frames, while these false positives can be distinguished considering the context information of the whole video.}
    \label{fig:motivation}
\end{figure}

Despite their effectiveness on still images, still-image object detection frameworks are not specifically designed for videos.
One key element of videos is temporal information, because locations and appearances of objects in videos should be temporally consistent, i.e. the detection results should not have dramatic changes over time in terms of both bounding box locations and detection confidences.
However, if still-image object detection frameworks are directly applied to videos, the detection confidences of an object show dramatic changes between adjacent frames and large long-term temporal variations, as shown by an example in Fig.~\ref{fig:motivation} (a).

One intuition to improve temporal consistency is to propagate detection results to neighbor frames to reduce sudden changes of detection results.
If an object exists in a certain frame, the adjacent frames are likely to contain the same object at neighboring locations with similar confidence.
In other words, detection results can be propagated to adjacent frames according to motion information so as to reduce missed detections. The resulted duplicate boxes can be easily removed by non-maximum suppression (NMS).

Another intuition to improve temporal consistency is to impose long-term constraints on the detection results.
As shown in Fig.~\ref{fig:motivation} (a), the detection scores of a sequence of bounding boxes of an object have large fluctuations over time.
These box sequences, or tubelets, can be generated by tracking and spatio-temporal object proposal algorithms \cite{Oneata:2014spatio}.
A tubelet can be treated as a unit to apply the long-term constraint.
Low detection confidence on some positive bounding boxes may result from moving blur, bad poses, or lack of enough training samples under particular poses.
Therefore, if most bounding boxes of a tubelet have high confidence detection scores, the low confidence scores at certain frames should be increased to enforce its long-term consistency.

Besides temporal information, contextual information is also a key element of videos compared with still images.
Although image context information has been investigated \cite{ouyang2015deepid} and incorporated into still-image detection frameworks, a video, as a collection of hundreds of images, has much richer contextual information.
As shown in Fig.~\ref{fig:motivation} (b), a small amount of frames in a video may have high confidence false positives on some background objects.
Contextual information within a single frame is sometimes not enough to distinguish these false positives.
However, considering the majority of high-confidence detection results within a video clip, the false positives can be treated as outliers and then their detection confidences can be suppressed.

The contribution of this works is three-folded.
1) We propose a deep learning framework that extends popular still-image detection frameworks (R-CNN and Faster R-CNN) to solve the problem of general object detection in videos by incorporating temporal and contextual information from tubelets. It is called T-CNN, i.e. tubelets with convolution neural network.
2) Temporal information is effectively incorporated into the proposed detection framework by locally propagating detection results across adjacent frames as well as globally revising detection confidences along tubelets generated from tracking algorithms.
3) Contextual information is utilized to suppress detection scores of low-confidence classes based on all detection results within a video clip.
This framework is responsible for winning the VID task with provided data and achieving the second place with external data in ImageNet Large-Scale Visual Recognition Challenge 2015 (ILSVRC2015).
Code is available at \url{https://github.com/myfavouritekk/T-CNN}.


\section{Related Work} 
\label{sec:related_work}
\textbf{Object detection from still images.}
State-of-the-art methods for detecting objects of general classes are mainly based on deep CNNs \cite{googlenet,sermanet2013overfeat,girshick2014rich,girshick2015fast,ren2015faster,erhan2014scalable,ouyang2015deepid,redmon2016you,gidaris2015object,girshick2015deformable,mrowca2015spatial,bell2015inside,yan2015object,he2016deep,wang2016dictionary}. Girshick et al. \cite{girshick2014rich} proposed a multi-stage pipeline called Regions with Convolutional Neural Networks
(R-CNN) for training deep CNNs to classify region proposals for object detection. It decomposed the detection problem into several stages including bounding-box proposal, CNN pre-training, CNN fine-tuning, SVM training, and bounding box regression. Such framework showed good performance and was widely adopted in other works. Szegedy et al. \cite{googlenet} proposed the GoogLeNet with a 22-layer structure and ``inception'' modules to replace the CNN in the R-CNN, which won the ILSVRC 2014 object detection task. Ouyang et al. \cite{ouyang2015deepid} proposed a deformation constrained pooling layer and a box pre-training strategy, which achieved an accuracy of $50.3\%$ on the ILSVRC 2014 test set. To accelerate the training of the R-CNN pipeline, Fast R-CNN \cite{girshick2015fast} was proposed, where each image patch was no longer wrapped to a fixed size before being fed into CNN. Instead, the corresponding features were cropped from the output feature maps of the last convolutional layer. In the Faster R-CNN pipeline \cite{ren2015faster}, the region proposals were generated by a Region Proposal Network (RPN), and the overall framework can thus be trained in an end-to-end manner. {He et al. proposed a novel Residual Neural Network (ResNet) \cite{he2016deep} based on residual blocks, which enables training very deep networks with over one hundred layers. Based on the Faster R-CNN framework, He et al. utilized ResNet to win the detection challenges in ImageNet 2015 and COCO 2015. The ResNet has later been applied to many other tasks and proven its effectiveness. Besides region based frameworks, some direct regression frameworks have also been proposed for object detection. YOLO \cite{redmon2016you} divided the image into even grids and simultaneously predicted the bounding boxes and classification scores. SSD \cite{liu2016ssd} generated multiple anchor boxes for each feature map location so as to predict bounding box regression and classification scores for bounding boxes with different scales and aspect ratios.} All these pipelines were for object detection from still images. When they are directly applied to videos in a frame-by-frame manner, they might miss some positive samples because objects might not be of their best poses at certain frames of videos.

{
\textbf{Object detection in videos.}
Since the introduction of ImageNet VID dataset in 2015, there has been multiple works that solve the video object detection problem. 
Han et al. \cite{han2016seq} proposed a sequence NMS method to associated still-image detections into sequences and apply the sequence-level NMS on the results. Weaker class scores are boosted by the detection on the same sequence. Galteri et al. \cite{galteri2017spatio} proposed a closed-loop framework to use object detection results on the previous frame to feed back to the proposal algorithm to improve window ranking. Kang et al. \cite{kang2017object} proposed a tubelet proposal network to efficiently generates hundreds of tubelet proposals simultaneously. 
}


\textbf{Object localization in videos.}
There have also been works on object localization and co-localization~\cite{Prest:2012learning,Papazoglou:2013fast,Joulin:2014efficient,Kwak:2015unsupervised,Maxime:2015is}.
Although such a task seems to be similar, the VID task we focus on is much more challenging.
There are crucial differences between the two problems.
1) The (co)localization problem only requires localizing one of the ground-truth objects of a known (weakly supervised setting) or unknown (unsupervised setting) class in each test frame. In contrast, the VID requires annotating every object from all target classes in each frame.
2) The CorLoc metric is used for localization problem, which is the percentage of test frames in which one of the ground-truth objects is correctly localized with intersection-over-union (IOU) $>0.5$. In VID, the mean average precision (Mean AP) is used for evaluation, which firstly calculates average precisions over all recall rates for each target class and then takes the mean value of all the average precisions.
3) The object (co)localization datasets \cite{Prest:2012learning,rubinstein2013unsupervised} contain annotations on only a few frames (\textit{e.g.}, $1$ frame for each training clip in the YouTubeObjects dataset), while VID dataset contains annotations for different classes on every training and test frames.
The much richer annotations not only enable investigating supervised deep learning methods which were not possible before, but also evaluates algorithm performances more precisely.
With the differences above, the VID task is much more challenging and promising for real-world applications. The previous works on object localization cannot be directly applied to the VID task, while the VID framework can be reduced to solve the object localization problem.

\textbf{Image classification.} The performance of image classification has been significantly improved during the past few years thanks to the large scale datasets \cite{deng2009imagenet} as well as novel deep neural networks and methods \cite{krizhevsky2012imagenet,googlenet,vgg2014simonyan,he2016deep,wang2016cost}. The models for object detection are commonly pre-trained on the ImageNet $1000$-class classification task. Batch normalization layer was proposed in \cite{ioffe2015batch} to reduce the statistical variations among mini batches and accelerate the training process. Simonyan et al. proposed a $19$-layer neural network with very small $3\times3$ convolution kernels in \cite{vgg2014simonyan}, which was proved effective in other related tasks such as detection \cite{girshick2015fast,ren2015faster}, action recognition \cite{Simonyan:2014twostream}, and semantic segmentation \cite{long2015fully}.

\textbf{Visual tracking.}
Object tracking has been studied for decades \cite{possegger2014occlusion,li2015reliable,hong2015multi,maggio2007adaptive,maggio2008efficient,cavallaro2005tracking,yang2009robust,li2017learning}. Recently, deep CNNs have been used for object tracking and achieved impressive tracking accuracy. Wang et al. \cite{wang2015visual} proposed to create an object-specific tracker by online selecting the most influential features from an ImageNet pre-trained CNN, which outperforms state-of-the-art trackers by a large margin.
Nam et al. \cite{nam2015learning} trained a multi-domain CNN for learning generic representations for tracking objects. When tracking a new target, a new network is created by combining the shared layers in the pre-trained CNN with a new binary classification layer, which is online updated.
Tracking is apparently different from VID, since it assumes the initial localization of an object in the first frame and it does not require predicting class labels.



\begin{figure*}[t]
    \centering
    \includegraphics[width=0.85\linewidth]{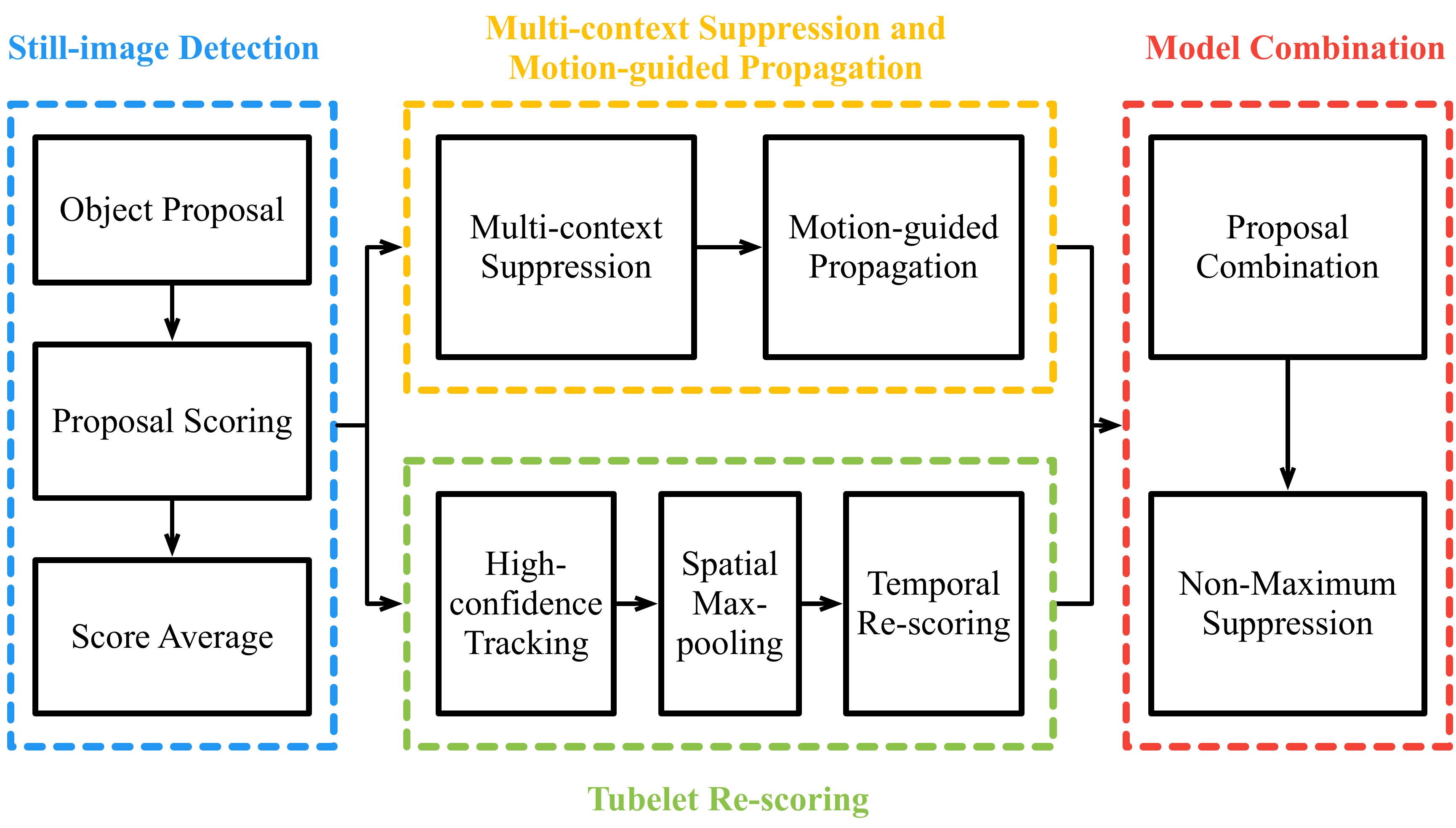}
    \caption{Our proposed T-CNN framework. The framework mainly consists of four components. 1) The still-image object detection component generates object region proposals in all the frames in a video clip and assigns each region proposal an initial detection score. 2) The multi-context suppression incorporates context information to suppress false positives and motion-guided propagation component utilizes motion information to propagate detection results to adjacent frames to reduce false negatives. 3) The tubelet re-scoring components utilizes tracking to obtain long bounding box sequences to enforce long-term temporal consistency of their detection scores. 4) The model combination component combines different groups of proposals and different models to generate the final results.}
    \label{fig:framework}
\end{figure*}

\section{Methods} 
\label{sec:methods}
In this section, we first introduce the VID task setting (Section~\ref{sub:task_setting}) and our overall framework (Section~\ref{sub:overview}).
Then each major component will be introduced in more details.
Section~\ref{sub:still_image} describes the settings of our still-image detectors. Section~\ref{sub:mcs_mgp} introduces how to utilize multi-context information to suppress false positive detections and utilize motion information to reduce false negatives. Global tubelet re-scoring is introduced in Section~\ref{sub:tubelet_re_scoring}.

\subsection{VID task setting} 
\label{sub:task_setting}
The ImageNet object detection from video (VID) task is similar to the object detection task (DET) in still images.
It contains $30$ classes to be detected, which are a subset of $200$ classes of the DET task. All classes are fully labeled in all the frames of each video clip.
For each video clip, algorithms need to produce a set of annotations $(f_i, c_i, s_i, b_i)$ of frame index $f_i$, class label $c_i$, confidence score $s_i$ and bounding box $b_i$.
The evaluation protocol for the VID task is the same as the DET task, i.e. we use the conventional mean average precision (Mean AP) on all classes as the evaluation metric.

\subsection{Framework overview} 
\label{sub:overview}
The proposed framework is shown in Fig.~\ref{fig:framework}.
It consists of four main components: 1) still-image detection, 2) multi-context suppression and motion-guided propagation, 3) temporal tubelet re-scoring, and 4) model combination.

\textbf{Still-image object detection.}
Our still-image object detectors adopt the DeepID-Net \cite{ouyang2015deepid} and CRAFT\cite{yang2015craft} frameworks and are trained with both ImageNet detection (DET) and video (VID) training datasets in ILSVRC 2015. DeepID-Net \cite{ouyang2015deepid} is an extension of R-CNN \cite{girshick2014rich} and CRAFT is an extension of Faster R-CNN \cite{ren2015faster}. Both of the two frameworks contain the steps of object region proposal and region proposal scoring.
The major difference is that in CRAFT (also Faster R-CNN), the proposal generation and classification are combined into a single end-to-end network.
Still-image object detectors are applied to individual frames.
The results from the two still-image object detection frameworks are treated separately for the remaining components in the proposed T-CNN framework.

\textbf{Multi-context suppression.}
This process first sorts all still-image detection scores within a video in descending orders. The classes with highly ranked detection scores are treated as high-confidence classes and the rest as low-confidence ones. The detection scores of low-confidence classes are suppressed to reduce false positives.

\textbf{Motion-guided Propagation.}
In still-image object detection, some objects may be missed in certain frames while detected in adjacent frames. Motion-guided propagation uses motion information such as optical flows to locally propagate detection results to adjacent frames to reduce false negatives.

\textbf{Temporal tubelet re-scoring.}
Starting from high-confidence detections by still-image detectors, we first run tracking algorithms to obtain sequences of bounding boxes, which we call tubelets.
Tubelets are then classified into positive and negative samples according to the statistics of their detection scores.
Positive scores are mapped to a higher range while negative ones to a lower range, thus increasing the score margins.

\textbf{Model combination.}
For each of the two groups of proposals from DeepID-Net and CRAFT, their detection results from both tubelet re-scoring and the motion-guided propagation are each min-max mapped to $[0,1]$ and combined by an NMS process with an IOU overlap $0.5$ to obtain the final results.

\subsection{Still-image object detectors} 
\label{sub:still_image}
Our still-image object detectors are adopted from DeepID-Net \cite{ouyang2015deepid} and CRAFT\cite{yang2015craft}.
The two detectors have different region proposal methods, pre-trained models and training strategies.

\subsubsection{DeepID-Net}
\textbf{Object region proposals.}
For DeepID-Net, the object region proposals are obtained by selective search (SS) \cite{uijlings2013selective} and Edge Boxes (EB) \cite{Zitnick:2014edgeboxes} with a cascaded selection process that eliminates easy false positive boxes using an ImageNet pre-trained AlexNet \cite{krizhevsky2012imagenet} model.
All proposal boxes are then labeled with $200$ ImageNet detection class scores by the pre-trained AlexNet.
The boxes whose maximum prediction scores of all $200$ classes are lower than a threshold are regarded as easy negative samples and are eliminated.
The process removes around $94\%$ of all proposal boxes while obtains a recall around $90\%$.

\textbf{Pre-trained models.}
ILSVRC 2015 has two tracks for each task.
1) For the provided data track, one can use data and annotations from all ILSVRC 2015 datasets including classification and localization (CLS), DET, VID and Places2.
2) For the external data track, one can use additional data and annotations.
For the provided data track, we pretrained VGG \cite{vgg2014simonyan} and GoogLeNet \cite{googlenet} with batch normalization (BN) \cite{ioffe2015batch} using the CLS $1000$-class data, while for the external data track, we used the ImageNet $3000$-class data. Pre-training is done at the object-level annotation as in \cite{ouyang2015deepid} instead of image-level annotation in R-CNN \cite{girshick2014rich}.

\textbf{Model finetuning and SVM training.}
Since the classes in VID are a subset of DET classes, the DET pretained networks and SVM can be directly applied to the VID task, with correct class index mapping.
However, due to the mismatch of the DET and VID data distributions and the unique statistics in videos, the DET-trained models may not be optimal for the VID task.
Therefore, we finetuned the networks and re-trained the $30$ SVMs with combination of DET and VID data.
Different combination configurations are investigated and a $2:1$ DET to VID data ratio achieves the best performance (see Section~\ref{sub:parameters}).

\textbf{Score average.}
Multiple CNN and SVM models are trained separately for the DeepID-Net framework, their results are averaged to generate the detection scores.
Such score averaging process is conducted in a greedy searching manner.
The best single model is first chosen. Then for each of the remaining models, its detection scores are averaged with those of the chosen model, and the model with best performance is chosen as the second chosen model. The process repeats until no significant improvement is observed.

\subsubsection{CRAFT}

CRAFT is an extension of Faster R-CNN. It contains the Region Proposal Network (RPN) stream to generated object proposals and the Fast-RCNN stream which further assigns a class (including background) score to each proposal.

\textbf{Object region proposals.}
In this framework, we use the enhanced version of Faster-RCNN by cascade RPN and cascade Fast-RCNN.
In our cascaded version of RPN, the proposals generated by the RPN are further fed into a object/background Fast-RCNN.
We find that it leads to a $93\%$ recall rate with about $100$ proposals per image.
In our cascade version of the Fast-RCNN, we further use a class-wise softmax loss as the cascaded step.
It is utilized for hard negative mining and leads to about $2\%$ improvement in mean AP.

\textbf{Pretrained models.}
Similar to the DeepID-Net setting, the pretrained models are the VGG and GoogLeNet with batch normalization. We only use the VGG in the RPN step and use both models in the later Fast-RCNN classification step.

\textbf{Score average.} The same greedy searching is conducted for model averaging as the DeepID-Net framework.


\begin{figure}[tb]
    \centering
    \includegraphics[width=0.95\linewidth]{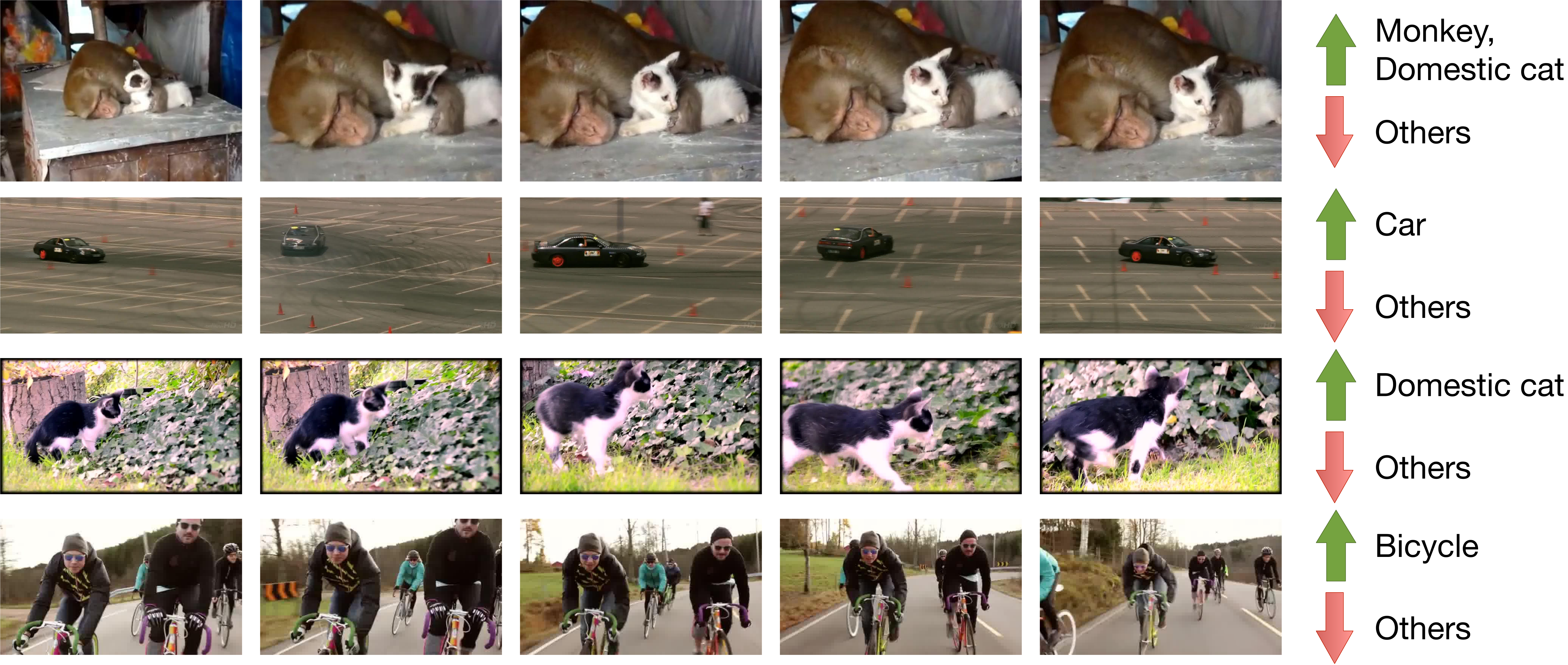}
    \caption{Multi-context suppression. For each video, the classes with top detection confidences are regarded as the high-confidence classes (green arrows) and others are regarded as low-confidence ones (red arrows). The detection scores of high-confidence classes are kept the same, while those of low-confidence ones are decreased to suppress false positives.}
    \label{fig:mcs}
\end{figure}

\subsection{Multi-context suppression (MCS) and motion-guided propagation (MGP)} 
\label{sub:mcs_mgp}
\textbf{Multi-context suppression (MCS).}
One limitation of directly applying still-image object detectors to videos is that they ignore the context information within a video clip.
The detection results in each frame of a video should be strongly correlated and we can use such property to suppress false positive detections.
We observed that although video snippets in the VID dataset may contain arbitrary number of classes, statistically each video usually contains only a few classes and co-existing classes have correlations.
Statistics of all detections within a video can therefore help distinguish false positives.

For example in Fig.~\ref{fig:mcs}, in some frames from a video clip, some false positive detections have very large detection scores.
Only using the context information within these frames cannot distinguish them from the positive samples.
However, considering the detection results on other frames, we can easily determine that the majority of high-confidence detections are other classes and these positive detections are outliers.

For each frame, we have about a few hundred region proposals, each of which has detection scores of $30$ classes.
For each video clip, we rank all detection scores on all boxes in a descending order.
The classes of detection scores beyond a threshold are regarded as high-confidence classes and the rest as low-confidence classes.
The detection scores of the high-confidence classes are kept the same, while those of the low-confidence classes are suppressed by subtracting a certain value.
The threshold and subtracted value are greedily searched on the validation set.
{
We observed that most videos contain only a small number of classes. We calculated the statistics of the number of classes on each video of the training set. The mean is $\mu=1.134$ and the standard deviation is $\sigma=0.356$. 
The probability of containing more than $2$ classes in a single video clip is therefore low. The MCS selects the top classes based on the ranking of box scores and punishes the unlikely classes because of the lower possibility of a large number of classes being in the same video.
}


\textbf{Motion-guided propagation (MGP).}
The multi-context suppression process can significantly reduce false positive detections, but cannot recover false negatives.
The false negatives are typically caused by several reasons.
1) There are no region proposals covering enough areas of the objects;
2) Due to bad pose or motion blur of an object, its detection scores are low.

These false negatives can be recovered by adding more detections from adjacent frames, because the adjacent frames are highly correlated, the detection results should also have high correlations both in spatial locations and detection scores.
For example, if an object is still or moves at a low speed, it should appear at similar locations in adjacent frames.
This inspires us to propagate boxes and their scores of each frame to its adjacent frame to augment detections and reduce false negatives.

\begin{figure}[tb]
    \centering
    \includegraphics[width=0.95\linewidth]{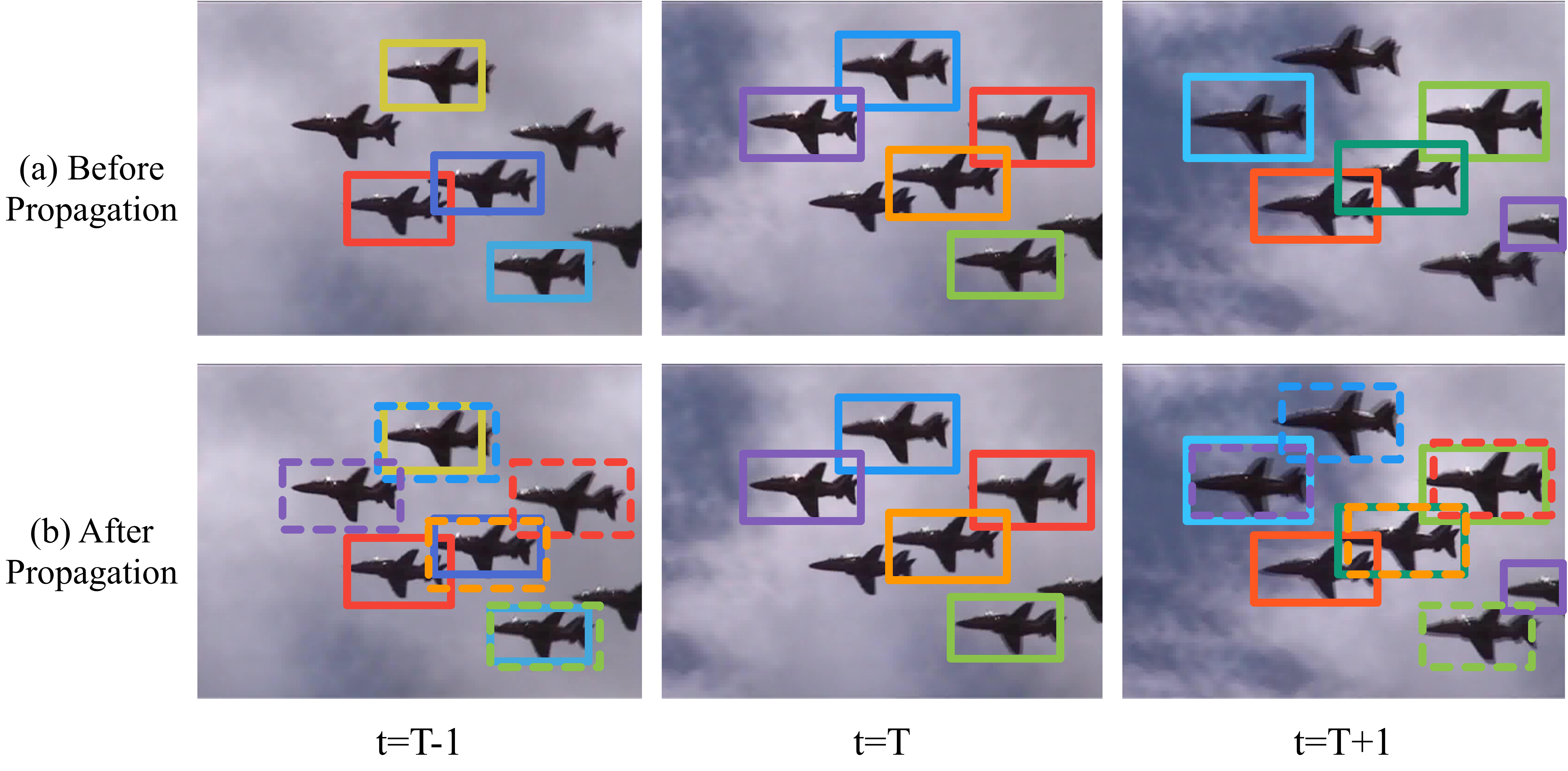}
    \caption{Motion-guided propagation. Before the propagation, some frames may contain false negatives (e.g. some airplanes are missing in (a)). Motion-guided propagation is to propagate detections to adjacent frames (e.g. from $t=T$ to $t=T-1$ and $t=T+1$) according to the mean optical flow vector of each detection bounding box. After propagation, fewer false negatives exist in (b).}
    \label{fig:mgp}
\end{figure}



{
If the baseline still-image detector is applied to the videos frame by frame, the false negatives may be cased by missing box proposals or low-detection scores. If we assume the probability of false negatives to be $p_n$, MGP of window $w$ propagates the results of $w$ neighboring frames to the current frame, which on average reduces the mis-detection rate to $p_n^w$ if the optical flow estimation is perfect. Although, long-range MGP is limited by the accuracy of optical flow estimation, the mis-detection rate can be generally reduced within small temporal windows, and the new false positives are unlikely to be added because of the final NMS operation.
}

We propose a motion-guided approach to propagate detection bounding boxes according to the motion information.
For each region proposal, we calculate the mean optical flow vector within the bounding box of the region proposal and propagate the box coordinates with same detection scores to adjacent frames according the mean flow vectors. An illustration example is shown in Fig.~\ref{fig:mgp}.


\subsection{Tubelet re-scoring} 
\label{sub:tubelet_re_scoring}
MGP generates short dense tubelets at every detection by our still-image detectors.
It significantly reduces false negatives but only incorporates short-term temporal constraints and consistency to the final detection results.
To enforce long-term temporal consistency of the results, we also need tubelets that span long periods of time.
Therefore, we use tracking algorithms to generate long tubelets and associate still-image object detections around tubelets.

As shown in Fig.~\ref{fig:framework}, the tubelet re-scoring includes three sub-steps: 1) high confidence tracking, 2) spatial max-pooling, and 3) tubelet classification.

\begin{figure}[tb]
    \centering
    \includegraphics[width=\linewidth]{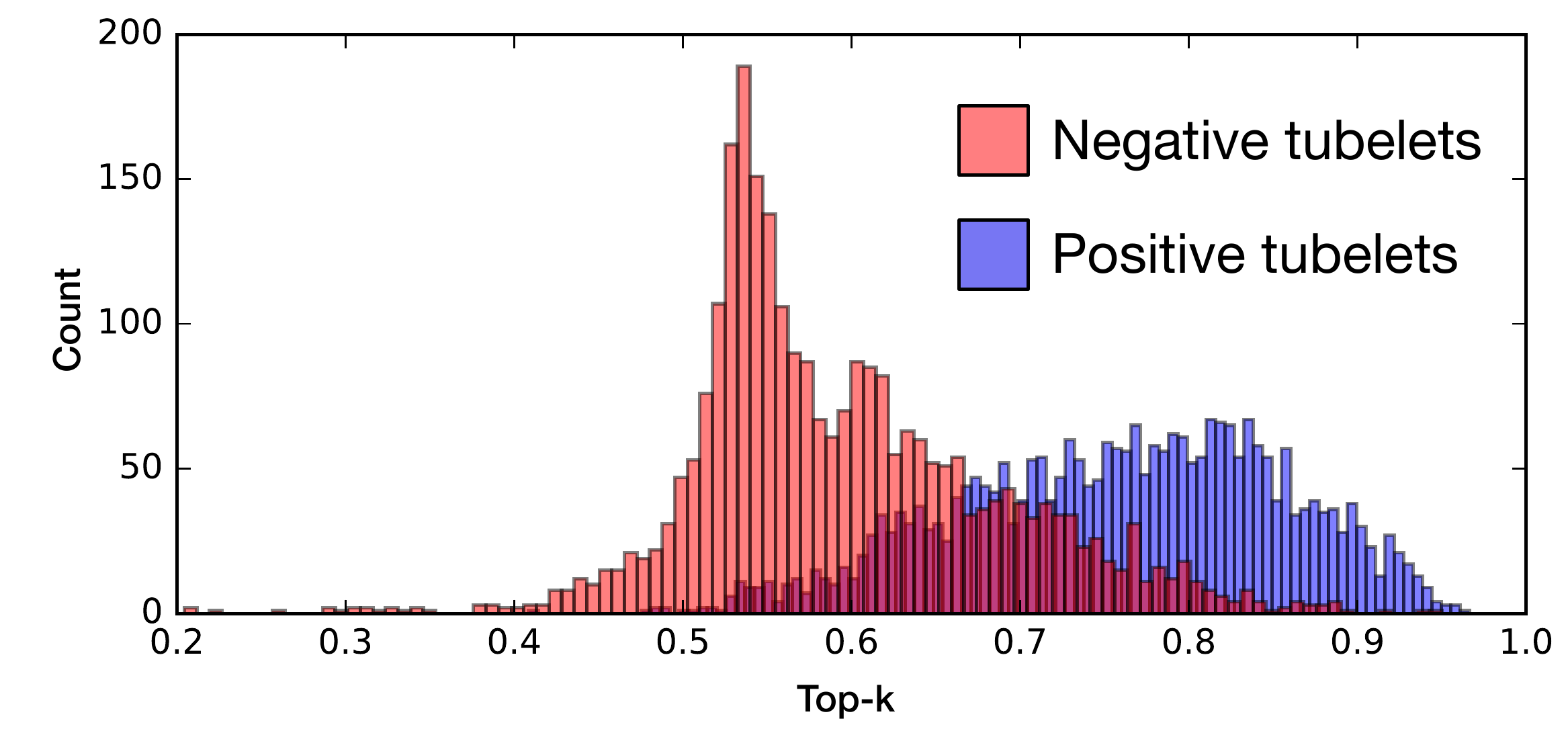}
    \caption{Tubelet classification. Tubelets obtained from tracking can be classified into positive and negative samples using statistics (e.g. top-k, mean, median) of the detection scores on the tubelets. Based on the statistics on the training set, a $1$-D Bayesian classifier is trained to classify the tubelets for re-scoring.}
    \label{fig:tubelet_rescoring}
\end{figure}

\textbf{High-confidence tracking.}
For each object class in a video clip, we track high-confidence detection proposals bidirectionally over the temporal dimension.
The tracker we choose is from \cite{wang2015visual}, which in our experiments shows robust performance to different object poses and scale changes.
The starting bounding boxes of tracking are called ``anchors'', which are determined as the most confident detections.
Starting from an anchor, we track biredictionally to obtain a complete tubelet.
As the tracking is conducted along the temporal dimension, the tracked box may drift to background or other objects, or may not adapt to the scale and pose changes of the target object. Therefore, we stop the tracking early when the tracking confidence is below a threshold (probability of $0.1$ in our experiments) to reduce false positive tubelets.
After obtaining a tubelet, a new anchor is selected from the remaining detections to start a new track.
Usually, high-confidence detections tend to cluster both spatially and temporally, and therefore directly tracking the next most confident detection tends to result in tubelets with large mutual overlaps on the same object.
To reduce the redundancy and cover as many objects as possible, we perform a suppression process similar to NMS.
Detections that have overlaps with the existing tracks beyond a certain threshold (IOU, i.e. Intersection of Union, $0.3$ in our experiment) will not be chosen as new anchors.
The tracking-suppression process performs iteratively until confidence values of all remaining detections are lower than a threshold.
For each video clip, such tracking process is performed for each of the $30$ VID classes.

\textbf{Spatial max-pooling.}
After tracking, for each class, we have tubelets with high-confidence anchors. A naive approach is to classify each bounding box on the tubelets using still-image object detectors.
Since the boxes from tracked tubelets and those from still-image object detectors have different statistics, when a still-image object detector is applied to a bounding box obtained from tracking, the detection score many not be accurate.
In addition, the tracked box locations may not be optimal due to the tracking failures. Therefore, the still-image detection scores on the tracked tubelets may not be reliable.

However, the detections spatially close to the tubelets can provide helpful information.
The spatial max-pooling process is to replace tubelet box proposals with detections of higher confidence by the still-image object detector.

For each tubelet box, we first obtain the detections from still-image object detectors that have overlaps with the box beyond a threshold (IOU $0.5$ in our setting).
Then only the detection with the maximum detection score is kept and used to replace the tracked bounding box.
This process is to simulate the conventional NMS process in object detection. If the tubelet box is indeed a positive box but with low detection score, this process can raise its detection score.
The higher the overlap threshold, the more confidence on the tubelet box.
In an extreme case when IOU = 1 is chosen as the threshold, we fully rely on the tubelet boxes while their surrounding boxes from still-image object detectors are not considered.

\textbf{Tubelet classification and rescoring.}
High-confidence tracking and spatial max-pooling generate long sparse tubelets that become candidates for temporal rescoring.
The main idea of temporal rescoring is to classify tubelets into positive and negative samples and map the detection scores into different ranges to increase the score margins.

Since the input only contains the original detection scores, the features for tubelet classification should also be simple.
We tried different statistics of tubelet detection scores such as mean, median and top-k (i.e. the kth largest detection score from a tubelet).
A Bayesian classifier is trained to classify the tubelets based on the statistics as shown in Fig.~\ref{fig:tubelet_rescoring}, and in our experiment, the top-k feature works best.

After classification, the detection scores of positive samples are min-max mapped to $[0.5,1]$, while negatives to $[0,0.5]$. Thus, the tubelet detection scores are globally changed so that the margins between positive and negative tubelets are increased.



\begin{figure*}[h]
    \centering
    \includegraphics[width=0.85\linewidth]{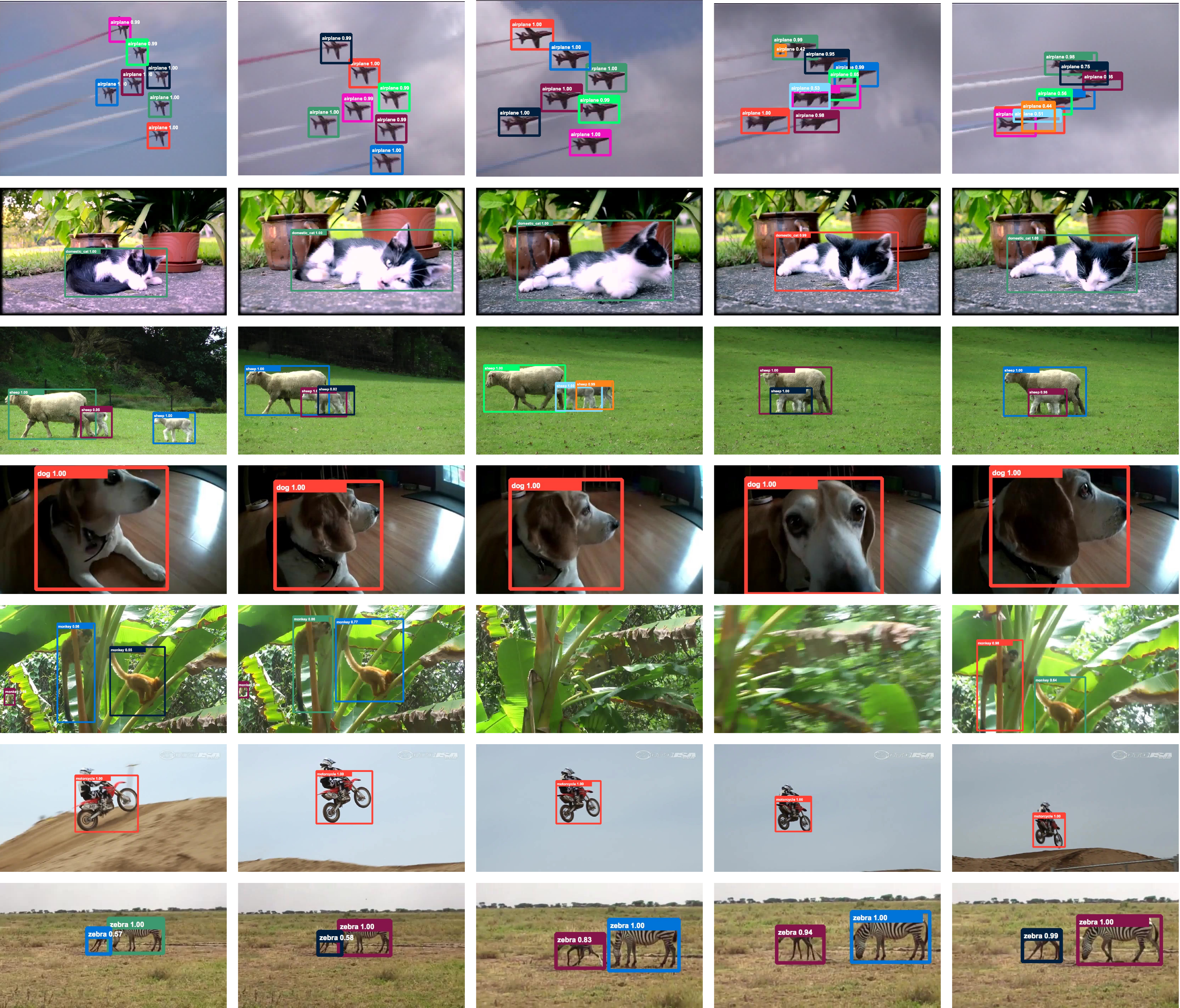}
    \caption{Qualitative results. The bounding boxes are tight to objects because of the combination of different region proposals. The detection results are consistent across adjacent frames thanks to motion-guided propagation and tracking. The false positives are much eliminated by multi-context suppression. (Different colors are used to mark bounding boxes in the same frame and do not represent tracking results)}
    \label{fig:qualitative}
\end{figure*}

\section{Experiments} 
\label{sec:experiments}
\subsection{Dataset} 
\label{sub:dataset}

\textbf{ImageNet VID Dataset}
The proposed framework is evaluated on the ImageNet VID dataset for object detection in videos.
1) The training set contains $3862$ fully-annotated video snippets ranging from $6$ frames to $5492$ frames per snippet.
2) The validation set contains $555$ fully-annotated video snippets ranging from $11$ frames to $2898$ frame per snippet.
3) The test set contains $937$ snippets and the ground truth annotation are not publicly available.
Since the official test server is primarily used for competition and has usage limitations, we primarily report the performances on the validation set as a common convention for object detection tasks. In the end, test results from top-ranked teams participated in ILSVRC 2015 are reported.

\textbf{YouTubeObjects (YTO) Dataset}
In addition to object detection on the ImageNet VID dataset, we also evaluated the object localization performance on the YTO dataset.
This dataset contains $10$ object classes with $870$ training clips and $334$ test clips.
There are several differences between ImageNet VID and YTO datasets.
1) VID dataset is fully annotated for all frames on both training and test sets, while YTO only contains bounding box annotations on very sparse frames ($4,306$ out of $521,831$ training frames and $1,781$ out of $198,321$ test frames).
2) The evaluation metric for VID is Mean AP and the system needs to detect every object in every frame. The YTO dataset, however, requires localizing objects of pre-known class for each video.


\subsection{Parameter Settings} 
\label{sub:parameters}
\textbf{Data configuration.}
We investigated the ratio of training data combination from the DET and VID training sets, and its influence on the still-image object detector DeepID-Net.
The best data configuration is then used for both DeepID-Net and CRAFT.
Because the VID training set has many more fully annotated frames than the DET training set, we kept all the DET images and sampled the training frames in VID for different combination ratios in order to training the still-image object detectors.

We investigated several training data configurations by finetuning a GoogLeNet with BN layers. From the TABLE~\ref{tab:data_config} and \ref{tab:data_config_craft}, we can see that the ratio of $2:1$ between DET and VID data has the best performance on the still-image detector DeepID-Net and CRAFT single models, therefore, we finetuned all of our models using this data configuration.

\begin{table}[!t]
    \caption{Performances of the still-image object detector DeepID-Net single model by using different finetuning data configurations on the initial validation set. The baseline DeepID-Net of only using the DET training data has the mean AP $49.8$.}
    \label{tab:data_config}
    \centering
    \normalsize

    \begin{tabular}{c|c|c|c|c|c}
    \hline
    \hline
    \textbf{DET:VID Ratio} & 1:0 & 3:1 & 2:1 & 1:1 & 1:3 \\
    \hline
        Mean AP / \% & 49.8 & 56.9 & \textbf{58.2} & 57.6 & 57.1\\
    \hline
    \hline
    \end{tabular}
\end{table}

\begin{table}[!t]
    \caption{Performances of the still-image object detector CRAFT single model by using different finetuning data configurations on the final validation set.}
    \label{tab:data_config_craft}
    \centering
    \normalsize

    \begin{tabular}{c|c|c}
    \hline
    \hline
    \textbf{DET:VID Ratio} & 0:1 & 2:1 \\
    \hline
        Mean AP / \% & 61.5 & 63.9\\
    \hline
    \hline
    \end{tabular}
\end{table}

\begin{table}[!t]
    \caption{Performances of different data configurations on the validation set for training SVMs in DeepID-Net. Baseline (the first column) only uses DET positive and negative samples and the result is a mean AP of $49.8$.}
    \label{tab:svm_config}
    \centering

    \begin{tabular}{c|c|c|c|c|c|c}
    \hline
    \hline
    DET Positive & \cmark & \cmark & \xmark & \xmark & \xmark & \cmark \\
    VID Positive & \xmark & \cmark & \cmark & \cmark & \cmark & \cmark \\
    DET Negative & \cmark & \cmark & \cmark & \cmark & \xmark & \cmark \\
    VID Negative & \xmark & \xmark & \xmark & \cmark & \cmark & \cmark \\
    \hline
    mean AP / \%  & 49.8   & 47.1   & 35.8   & 51.6   & 52.3   & \textbf{53.7}   \\
    \hline
    \hline
    \end{tabular}
\end{table}

\begin{table}[!h]
    \caption{Performances on the validation set by different temporal window sizes of MGP.}
    \label{tab:mgp}
    \centering

    \begin{tabular}{c|c|c|c|c|c}
    \hline
    \hline
    \multirow{2}{*}{Methods} & \multicolumn{5}{c}{Temporal window size} \\
    \cline{2-6}
    & 1 (baseline) & 3 & 5 & 7 & 9 \\
    \hline
    Duplicate & \multirow{2}{*}{70.7} & 71.7 & 72.1 & 71.5 & - \\
    \cline{1-1}
    \cline{3-6}
    Motion-guided &  & 71.7 & \textbf{73.0} & \textbf{73.0} & 72.6 \\
    \hline
    \hline
    \end{tabular}
\end{table}

\begin{table*}[!t]
    \caption{Performances of individual components, frameworks and our overall system.}
    \label{tab:component_results}
    \centering
    \small{

    \begin{tabular}{c|c|c|c|c|c|c|c|c|c}
    \hline
    \hline
    \multirow{2}{*}{Data} & \multirow{2}{*}{Model} & \multirow{2}{*}{Still-image} & \multirow{2}{*}{{+MGP}} & {+MGP} & {+MGP+MCS} & Model & \multirow{2}{*}{Test Set} & Rank in  & \multirow{2}{*}{\#win} \\
    & & & & {+MCS} & {+Rescoring} & Combination & & ILSVRC2015 & \\
    \hline
    \multirow{2}{*}{Provided} & CRAFT\cite{yang2015craft} & 67.7 & {68.7} & {72.4} & {73.6} & \multirow{2}{*}{73.8} & \multirow{2}{*}{67.8} & \multirow{2}{*}{\#1} & \multirow{2}{*}{28/30} \\
    \cline{2-6}
    & DeepID-net \cite{ouyang2015deepid} & 65.8 & {68.3} & {72.1} & {72.5} & & & &\\
    \hline
    \multirow{2}{*}{Additional} & CRAFT\cite{yang2015craft} & 69.5 & {70.3} & {74.1} & {75.0} & \multirow{2}{*}{77.0} & \multirow{2}{*}{69.7} & \multirow{2}{*}{\#2} & \multirow{2}{*}{11/30} \\
    \cline{2-6}
    & DeepID-net \cite{ouyang2015deepid} & 70.7 & {72.7} & {74.9} & {75.4} & & & &\\
    \hline
    \hline
    \end{tabular}
    }
\end{table*}

\begin{table*}[!t]
\caption{Performaces of our final models on the validation set.}
\label{tab:meanap}
\centering
\scriptsize{
\begin{tabular}{p{1.3cm}|p{0.5cm}p{0.5cm}p{0.5cm}p{0.5cm}p{0.5cm}p{0.5cm}p{0.5cm}p{0.5cm}p{0.5cm}p{0.5cm}p{0.5cm}p{0.5cm}p{0.5cm}p{0.5cm}p{0.5cm}p{0.5cm}}
\hline
\hline
Data & \rotatebox{60}{airplane} &  \rotatebox{60}{antelope} &  \rotatebox{60}{bear} &  \rotatebox{60}{bicycle} &  \rotatebox{60}{bird} &  \rotatebox{60}{bus} &  \rotatebox{60}{car} &  \rotatebox{60}{cattle} &  \rotatebox{60}{dog} &  \rotatebox{60}{d\_cat} &  \rotatebox{60}{elephant} &  \rotatebox{60}{fox} &  \rotatebox{60}{g\_panda} &  \rotatebox{60}{hamster} &  \rotatebox{60}{horse} &  \rotatebox{60}{lion} \\
\hline
Provided & 83.70 &  85.70 &  84.40 &  74.50 &  73.80 &  75.70 &  57.10 &  58.70 &  72.30 &  69.20 &  80.20 &  83.40 &  80.50 &  93.10 &  84.20 &  67.80 \\
\hline
Additional & 85.90 &  86.90 &  87.80 &  77.90 &  74.70 &  77.50 &  59.00 &  70.90 &  74.40 &  79.60 &  80.40 &  83.90 &  82.40 &  95.80 &  87.80 &  64.10 \\
\hline
\hline
Data & \rotatebox{60}{lizard} &  \rotatebox{60}{monkey} &  \rotatebox{60}{motorcycle} &  \rotatebox{60}{rabbit} &  \rotatebox{60}{r\_panda} &  \rotatebox{60}{sheep} &  \rotatebox{60}{snake} &  \rotatebox{60}{squirrel} &  \rotatebox{60}{tiger} &  \rotatebox{60}{train} &  \rotatebox{60}{turtle} &  \rotatebox{60}{watercraft} &  \rotatebox{60}{whale} &  \rotatebox{60}{zebra} &  \rotatebox{60}{mean AP} & \rotatebox{60}{\#win} \\
\hline
Provided & 80.30 &  54.80 &  80.60 &  63.70 &  85.70 &  60.50 &  72.90 &  52.70 &  89.70 &  81.30 &  73.70 &  69.50 &  33.50 &  90.20 &  \textbf{73.80} & 28/30 \\
\hline
Additional & 82.90 &  57.20 &  81.60 &  77.50 &  79.70 &  68.00 &  77.70 &  58.30 &  90.10 &  85.30 &  75.90 &  71.20 &  43.20 &  91.70 &  \textbf{77.00} & 11/30 \\
\hline
\hline
\end{tabular}
}
\end{table*}

In addition to model finetuning, we also investigated the data configurations for training the SVMs in DeepID-Net.
The performances are shown in TABLE~\ref{tab:svm_config}, which show that using positive and negative samples from both DET and VID data leads to the best performance.


Because of the redundancy among video frames, we also sampled the video frames by a factor of $2$ during testing and applied the still-image detectors to the remaining frames. The MCS, MGP and re-scoring steps in Section~\ref{sub:mcs_mgp} and \ref{sub:tubelet_re_scoring} are then conducted.
The detection boxes on the unsampled frames are generated by interpolation and MGP.
We did not observe significant performance differences with frame sampling on the validation set.

To conclude, we sampled VID frames to half the amount of DET images and combined the samples to finetune the CNN models in both DeepID-Net and CRAFT. Positive and negative samples from both DET and VID images are used to train SVMs in DeepID-Net.

\textbf{Hyperparameter settings.}
For motion-guided propagations, as described in Section~\ref{sub:mcs_mgp}, TABLE~\ref{tab:mgp} shows the performances of different propagation window sizes.
Compared to directly duplicating boxes to adjacent frames without changing their locations according to optical flow vectors, MGP has better performances with the same propagation windows, which proves that MGP generates detections with more accurate locations. $7$ frames ($3$ frames forward and $3$ backward) are empirically set as the window size.

In multi-context suppression, classes in the top $0.0003$ of all the bounding boxes in a video are regarded as high-confidence classes and the detection scores for both frameworks are subtracted by $0.4$.
Those hyperparameters are greedily searched in the validation set.

\textbf{Network configurations.}
The models in DeepID-Net and CRAFT are mainly based on GoogLeNet with batch-normalization layers and VGG models.
The techniques of multi-scale \cite{zeng2015window} and multi-region \cite{gidaris2015object} are used to further increase the number of models for score averaging in the still-image object detection shown in Fig.~\ref{fig:framework}.
The performance of a baseline DeepID-Net trained on ImageNet DET task can be increased from $49.8$ to $70.7$ with all the above-mentioned techniques (data configuration for finetuning, multi-scale, multi-region, score average, etc.).




\section{Results} 
\label{sec:results}

\subsection{Results on the ImageNet VID dataset} 
\label{sub:imagenet_vid_res}
\textbf{Qualitative results.} 
Some qualitative results of our proposed framework are shown in Fig.~\ref{fig:qualitative}.
From the figure, we can see the following characteristics of our proposed framework.
1) The bounding boxes are very tight to the objects, which results from the high-quality bonding box proposals combined from Selective Search, Edge Boxes and Region-proposal Networks.
2) The detections are consistent across adjacent frames without obvious false negatives thanks to the motion-guided propagation and tracking.
3) There are no obvious false positives even though the scenes may be complex (e.g. cases in the third row), because the multi-context information is used to suppress their scores.

\textbf{Quantitative results} 
The component analysis of our framework on both provided-data and additional-data tracks are shown in TABLE~\ref{tab:component_results}. The results are obtained from the validation set. From the table, we can see that the still-image object detectors obtain about $65-70\%$ Mean AP. Adding temporal and contextual information through MCS, MGP and tubelet re-scoring significantly improves the results by up to $6.7$ percents. {As shown in TABLE~\ref{tab:component_results}, the MGP generally has $0.8$-$2.5\%$ improvement, MCS has $2.2$-$3.8\%$ improvement, and tubelet rescoring has about $0.4\%$-$1.2\%$ improvement for different models.} The final model combination process further improves the performance.

Overall, our framework ranks $1$st on the provided-data track in ILSVRC2015 winning $28$ classes out of $30$ and $2$nd on the additonal-data track winning $11$ classes. The detailed AP lists of the submitted models on the validation set are shown in TABLE~\ref{tab:meanap}. The final results of our team and other top-ranked teams on the test data are shown in TABLE~\ref{tab:offcial_results}.

\begin{table}[!h]
    \caption{Performance comparison with other teams on ILSVRC2015 VID test set with provided data (sorted by mean AP, the best model is chosen for each team).}
    \label{tab:offcial_results}
    \centering
    \normalsize

    \begin{tabular}{c|c|c|c}
    \hline
    \hline
    \textbf{Rank} & \textbf{Team name} & \textbf{mean AP} & \textbf{\#win} \\
    \hline
    \hline
    \textbf{1} & \textbf{CUVideo (Ours)} & \textbf{67.82} & \textbf{28} \\
    \hline
    2 & ITLab VID - Inha & 51.50 & 0 \\
    \hline
    3 & UIUC-IFP\cite{han2016seq} & 48.72 & 0 \\
    \hline
    4 & Trimps-Soushen & 46.12 & 0 \\
    \hline
    5 & 1-HKUST & 42.11 & 0 \\
    \hline
    6 & HiVision & 37.52 & 0 \\
    \hline
    7 & RUC\_BDAI & 35.97 & 2 \\
    \hline
    \hline
    \end{tabular}
\end{table}

{We also used the same framework (without tubelet re-scoring due to time limit) to participate in ILSVRC2016 VID challenge with provided training data. The results are shown in TABLE~\ref{tab:imagenet2016}. The final ensemble model for the challenge consists of 4 models, including 1) a ResNet-101 model, 2) a ResNet-269 model, 3) a ResNet-269 model with gated bi-directional (GBD) CNN \cite{zeng2016gated} multi-context structures, and 4) an Inception-ResNet-v2 model \cite{szegedy2016inception}. From the table we can see that compared to the results in ILSVRC2015, the 4-model ensemble has about $12.5\%$ ($73.8$ to $86.3$) improvement on validation set and $9.0\%$ ($67.8$ to $76.8$) improvement on the test set, thanks to the new architectures such as ResNet, GBD-net and Inception-ResNet. The MCS+MGP technique still has about $3.3\%$ improvement given the much improved baseline. Overall, the proposed framework ranked No. 2 in ILSVRC 2016 VID challenge with provided training data.}

{
\begin{table}[h]
    {\caption{4-model ensemble in ILSVRC 2016 challenge.}
    \label{tab:imagenet2016}}
    \centering
    \normalsize
{
    \begin{tabular}{c|c|c|c|c}
    \hline
    \hline
    \multirow{2}{*}{Still-image} & \multirow{2}{*}{+MCS} & \multirow{2}{*}{+MGP} & test & Rank in\\
    & & & set & ILSVRC2016\\
    \hline
    83.0 & 85.9 & 86.3 & 76.8 & \#2 \\
    \hline
    \hline
    \end{tabular}}
\end{table}
}


{
\subsection{Results analysis} 
\label{sub:results_analysis}

\noindent\textbf{MGP in moving-camera sequences.} We evaluated the MGP technique in moving camera sequences. The validation videos are manually labeled into two subsets based on whether the video contains significant camera motion. In fact, the ImageNet VID dataset mainly consists of online videos, most of which are shot with hand-held cameras. The \textit{moving-camera} subset contains $401$ out of $555$ total validation videos. The mean APs of the DeepID-net model are shown in TABLE~\ref{tab:moving_camera}, from which we can see the improvement of MGP is consistent on the moving-camera subset.
One interesting observation is that the MGP has better improvement on the moving-camera subset than the full set. The reason is that usually for moving-camera sequences, the camera moves to focus on the objects of interest. Therefore, although the background is moving, the objects are relatively stable in the scenes.

\begin{table}[h]
    {\caption{MGP of DeepID-net on moving-camera subset.}
    \label{tab:moving_camera}}
    \centering
    \normalsize
{
    \begin{tabular}{c|c|c}
    \hline
    \hline
    \textbf{Set} & \textbf{still-image} & \textbf{+MGP} \\
    \hline
    \textbf{Full validation} & 70.7 & 72.7 \\
    \hline
    \textbf{Moving-camera subset} & 72.3 & 74.5 \\
    \hline
    \hline
    \end{tabular}}
\end{table}

\noindent\textbf{Time efficiency.} The proposed techniques add an limited amount of computation time to the still-image baseline. For MGP, the main additional computation is the optical flow generation. We adopted the Gunnar Farneback's algorithm in OpenCV and the speed is about $10$ fps. For MCS, the parameters are greedily searched using the detection scores, which requires little additional computation. For tubelet re-scoring, the bottleneck is the single-object tracking. The tracker that we use is a deep neural network tracker of $0.5$ fps. Since we only generates about $10$ tubelets for each video, it takes about $3$ days for completing the tracking for both training and validation set, which consists of $4417$ videos. In total, with a $8$-GPU server, it takes about $1$ week to pre-train a model on ImageNet classification task, $2$ days for fine-tuning on DET dataset, and $2$ days for fine-tuning on the VID dataset. The test time takes about $4$ days in total ($1$ day for generating still-image detections, $3$ days for incorporating temporal and context information). Considering there are a total of $3862$ training video, $555$ validation videos and $937$ test videos in the dataset with over $1$ million frames, the computational cost is acceptable.
}


\subsection{Localication on YouTubeObjects (YTO) Dataset} 
\label{sub:yto_res}
In addition to the detection results on the ImageNet VID dataset, we also evaluated our framework on the YouTubeObjects dataset for the object localization task.
Different from the object detection task, object localization requires the system to localize the objects of certain categories on certain video frames.
Since the classes of the YTO dataset is a subset of ImageNet VID dataset, the system is directly evaluated on the YTO dataset without any fine-tuning.
For each test frame, we only keep the detection result with the maximum detection score.
The evaluation metric is CorLoc \cite{Deselaers:2010localizing} as a common practice on YTO.
The CorLoc metric is the percentage of images in which the system correctly localizes one object of the target class with an overlap of IOU $>0.5$.

\begin{table*}[h]
    \caption{Object localization performances on the YTO dataset}
    \label{tab:yto_scores}
    \centering
    \begin{tabular}{l|p{0.5cm}p{0.5cm}p{0.5cm}p{0.5cm}p{0.5cm}p{0.5cm}p{0.5cm}p{0.5cm}p{0.5cm}p{0.5cm}|c}
    \hline
    \hline
    Method & aero & bird & boat & car & cat & cow & dog & horse & mbike & train & Avg. \\
    \hline
    \hline
    Prest \textit{et al.} \cite{Prest:2012learning} & 51.7 & 17.5 & 34.4 & 34.7 & 22.3 & 17.9 & 13.5 & 26.7 & 41.2 & 25.0 & 28.5 \\
    Joulin \textit{et al.} \cite{Joulin:2014efficient} & 25.1 & 31.2 & 27.8 & 38.5 & 41.2 & 28.4 & 33.9 & 35.6 & 23.1 & 25.0 & 31.0 \\
    Kwak \textit{et al.} \cite{Kwak:2015unsupervised} & 56.5 & 66.4 & 58.0 & 76.8 & 39.9 & 69.3 & 50.4 & 56.3 & 53.0 & 31.0 & 55.7 \\
    Kang \textit{et al.} \cite{Kang:2016object} & 94.1 & 69.7 & 88.2 & 79.3 & 76.6 & 18.6 & 89.6 & 89.0 & 87.3 & 75.3 & 76.8 \\
    \hline
    Baseline & 91.2 & 98.1 & 85.4 & 95.0 & 92.2 & 100.0 & 96.3 & 92.8 & 91.1 & 83.0 & 92.5\\
    Baseline + MGP (window 3) & 91.8 & 98.7 & 85.4 & 95.0 & 92.2 & 100.0 & 95.7 & 93.4 & 93.9 & 84.2 & \textbf{93.0}\\
    Baseline + MGP (window 5) & 91.8 & 98.1 & 86.1 & 94.2 & 90.1 & 99.3 & 93.9 & 93.4 & 92.7 & 86.1 & 92.6\\
    \hline
    \hline
    \end{tabular}
\end{table*}

The localization performances of our system and some of the state-of-the-art results on the YTO dataset are shown in TABLE~\ref{tab:yto_scores}.
The ``baseline'' network we used for this task is the best CRAFT single model finetuned with DET and VID data of ratio $2:1$. Because only the detection with the highest score in each frame is kept, we could only apply MGP post-processing to the baseline network to improve its performance (denoted as ``baseline + MGP'').
From the table we can see that our proposed framework outperforms previous methods with large margin. Even compared to the state-of-the-art method \cite{Kang:2016object}, our system still obtains about $15$ percents improvement.
We also compared with MGP post-processing with window size of $3$ and $5$. From the table,  we can see that the window size $3$ improves the performance about $0.5$ percent, while increasing the window size to $5$ causes performance drop.


\section{Conclusion} 
\label{sec:conclusion}
In this work, we propose a deep learning framework that incorporates temporal and contextual information into object detection in videos.
This framework achieved the state-of-the-art performance on the ImageNet object detection from video task and won the corresponding VID challenge with provided data in ILSVRC2015.
The component analysis is investigated and discussed in details.
Code is publicly available.

The VID task is still new and under-explored.
Our proposed framework is based on the popular still-image object detection frameworks and adds important components specifically designed for videos.
We believe that the knowledge of these components can be further incorporated in to end-to-end systems and is our future research direction.



%





\ifCLASSOPTIONcaptionsoff
  \newpage
\fi



%




%
\vspace{-1.5cm}
\begin{IEEEbiography}[{\includegraphics[width=1in,height=1.25in,clip,keepaspectratio]{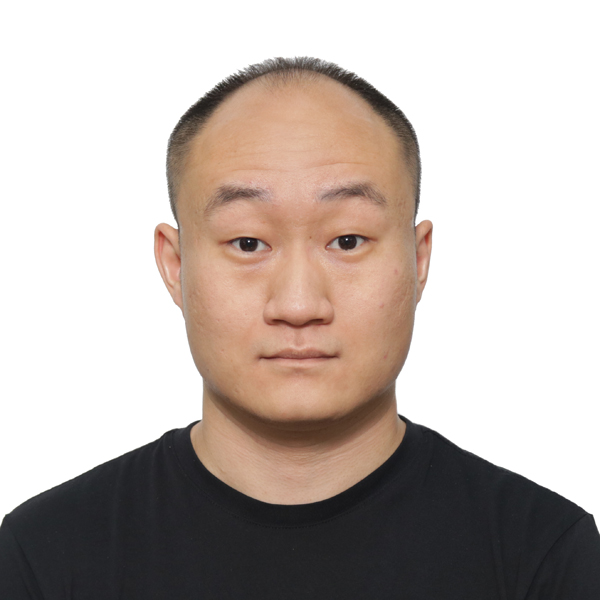}}]{Kai Kang} graduated from School of the Gifted Young at University of Science and Technology of China (USTC) in 2013 with a BS (Hons) degree in Optics. He is currently a PhD candidate in Electronic Engineering at the Chinese University of Hong Kong (CUHK). His research interests include computer vision, video analysis and deep learning.
\end{IEEEbiography}
\vspace{-1.5cm}
\begin{IEEEbiography}[{\includegraphics[width=1in,height=1.25in,clip,keepaspectratio]{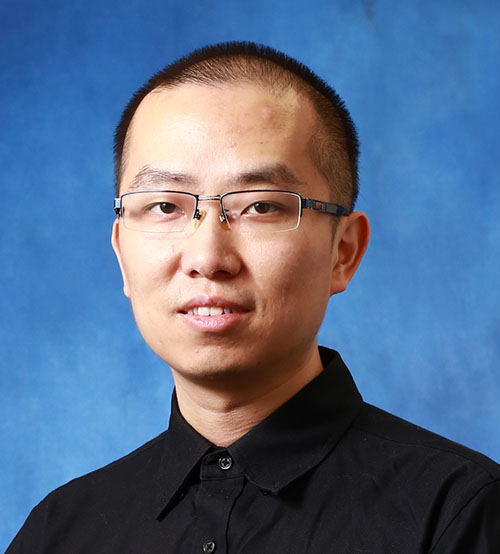}}]{Hongsheng Li} received the bachelors degree in automation from East China University of Science and Technology, and the masters and doctorate degrees in computer science from Lehigh University, Pennsylvania, in 2006, 2010, and 2012, respectively. He is a research assistant professor in the Department of Electronic Engineering at the Chinese University of Hong Kong. His research interests include computer vision, medical image analysis and machine learning.
\end{IEEEbiography}
\vspace{-1.5cm}
\begin{IEEEbiography}[{\includegraphics[width=1in,height=1.25in,clip,keepaspectratio]{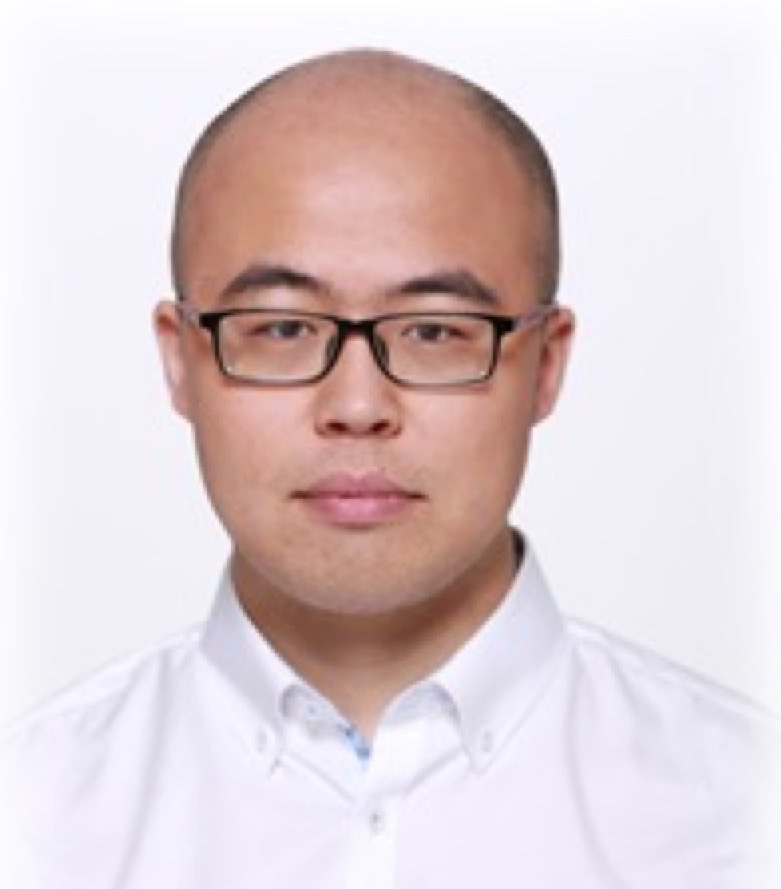}}]{Junjie Yan} is the R\&D Director at SenseTime Group Limited. In SenseTime, he leads the R\&D in detection, tracking, recognition, and video intelligence. He graduated in Jul. 2015 with a PhD degree at National Laboratory of Pattern Recognition, Chinese Academy of Sciences, where he mainly worked on object detection and face recognition.
\end{IEEEbiography}
\vspace{-1.5cm}
\begin{IEEEbiography}[{\includegraphics[width=1in,height=1.25in,clip,keepaspectratio]{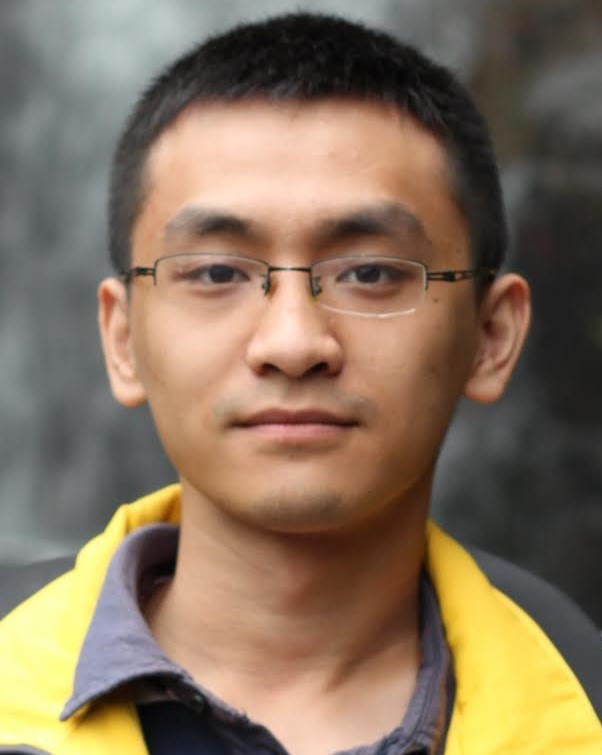}}]{Xingyu Zeng} received the PhD degree in the Department of Electronic Engineering from Chinese University of Hong Kong in 2016 and the BS degree in electronic engineering and information science from the University of Science and Technology, China, in 2011. He is currently working in Sensetime Group Limited. His research interests include computer vision and deep learning. 
\end{IEEEbiography}
\vspace{-1.5cm}
\begin{IEEEbiography}[{\includegraphics[width=1in,height=1.25in,clip,keepaspectratio]{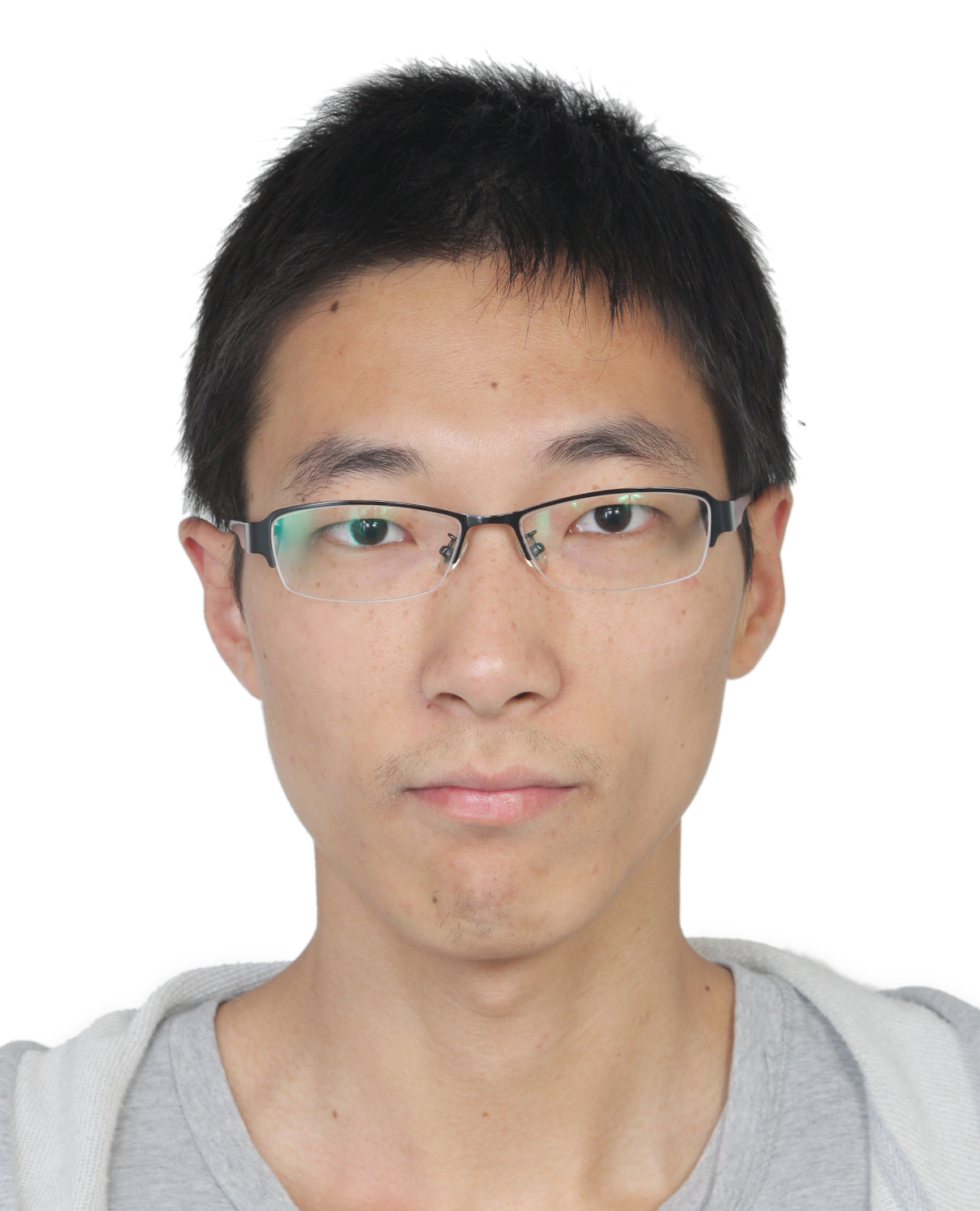}}]{Bin Yang} received the B.Eng. degree in Electronics and Information Engineering from China Agricultural University in 2014. He is currently pursuing the M.Sc. degree with the Computer Science Department, University of Toronto. His research interests include computer vision and machine learning. 
\end{IEEEbiography}
\vspace{-1.5cm}
\begin{IEEEbiography}[{\includegraphics[width=1in,height=1.25in,clip,keepaspectratio]{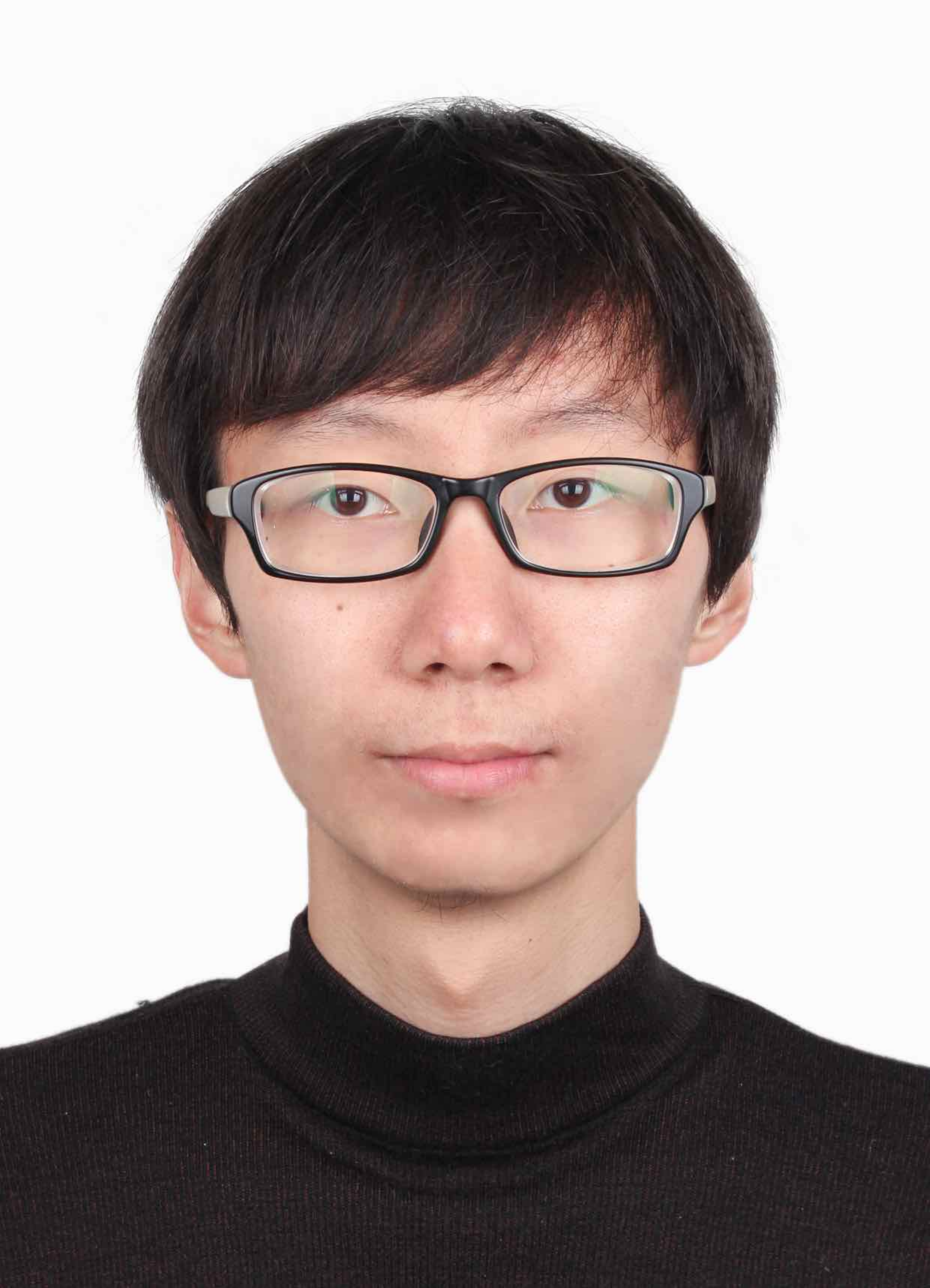}}]{Tong Xiao} is a Ph.D. candidate in electronic engineering at The Chinese University of Hong Kong, advised by Prof. Xiaogang Wang. Tong received Bachelor of Engineering in computer science from Tsinghua University. His research interests include computer vision and deep learning, especially learning better feature representations for human identification. He also served as a reviewer of top-tier computer vision conferences and journals, including CVPR, ICCV, ECCV, TIP, TNNLS, and TCSVT.
\end{IEEEbiography}
\vspace{-1.5cm}
\begin{IEEEbiography}[{\includegraphics[width=1in,height=1.25in,clip,keepaspectratio]{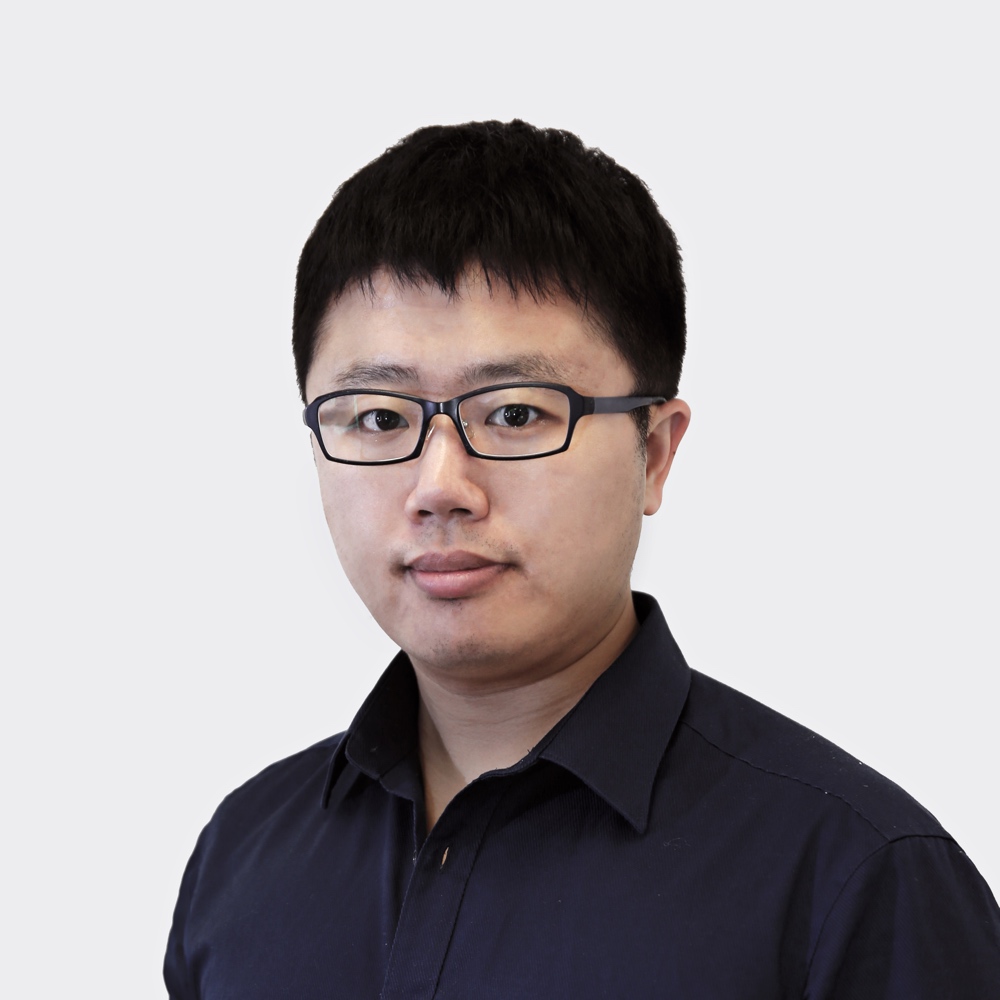}}]{Cong Zhang} received the B.S. degree from Shanghai Jiao Tong University, Shanghai in 2009 from Department of Electronic Engineering. He is currently pursuing the Ph.D. degree at the Institute of Image Communication and Information Processing, Shanghai Jiao Tong University. His current research interests include crowd understanding, crowd behavior detection, and machine learning.
\end{IEEEbiography}
\vspace{-1.5cm}
\begin{IEEEbiography}[{\includegraphics[width=1in,height=1.25in,clip,keepaspectratio]{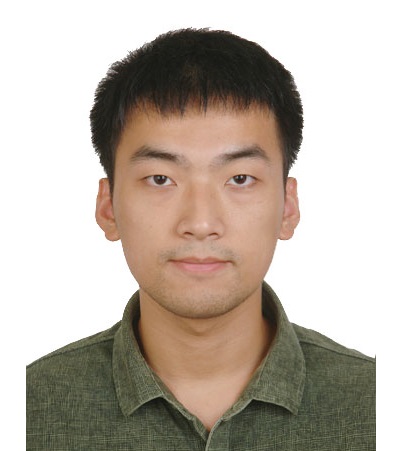}}]{Zhe Wang} received the BEng from the Department of Optical Engineering of Zhejiang University, China, in 2012. He is currently working toward a PhD degree in the Department of Electronic Engineering, the Chinese University of Hong Kong. His research interest is focused on deep learning and its applications, especially on medical imaging, segmentation and detection.
\end{IEEEbiography}
\vspace{-1.5cm}
\begin{IEEEbiography}[{\includegraphics[width=1in,height=1.25in,clip,keepaspectratio]{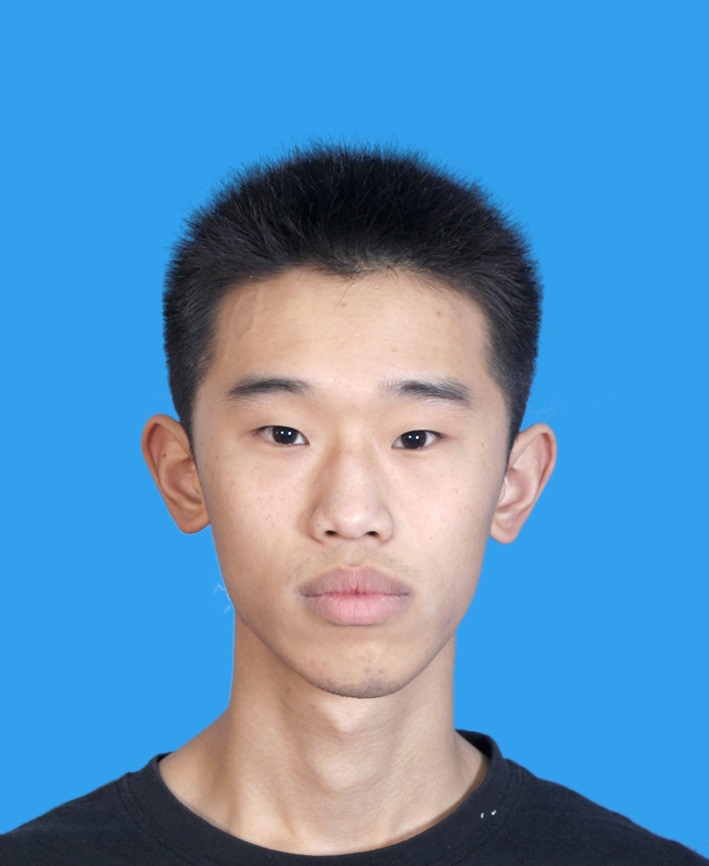}}]{Ruohui Wang} received his B.Eng. degree from Department of Electronic Engineering at Tsinghua University in 2013. He is currently a Ph.D. student in Department of Information Engineering at the Chinese University of Hong Kong. His research interests include machine learning and distributed computing.
\end{IEEEbiography}
\vspace{-1.5cm}
\begin{IEEEbiography}[{\includegraphics[width=1in,height=1.25in,clip,keepaspectratio]{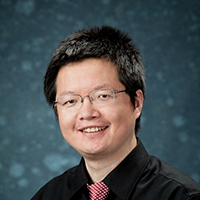}}]{Xiaogang Wang} received the BS degree from the Special Class for Gifted Young at University of Science and Technology of China in electrical engineering and information science in 2001, and the MPhil degree from Chinese University of Hong Kong in 2004. He received the PhD degree in computer science from the Massachusetts Institute of Technology. He is currently an associate professor in the Department of Electronic Engineering at the Chinese University of Hong Kong. His research interests include computer vision and machine learning.
\end{IEEEbiography}
\vspace{-1.5cm}
\begin{IEEEbiography}[{\includegraphics[width=1in,height=1.25in,clip,keepaspectratio]{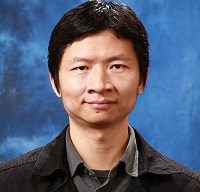}}]{Wanli Ouyang} received the PhD degree in the Department of Electronic Engineering, The Chinese University of Hong Kong. Since 2017, he is a senior lecturer at the University of Sydney. His research interests include image processing, computer vision and pattern recognition.
\end{IEEEbiography}







\end{document}